\documentclass[conference]{IEEEtran}
\IEEEoverridecommandlockouts
\usepackage[utf8]{inputenc}
\usepackage[T1]{fontenc}
\usepackage{newunicodechar}
\newunicodechar{✓}{\checkmark}
\usepackage{cite}
\usepackage{amsmath,amssymb,amsfonts}
\usepackage{algorithmic}
\usepackage{graphicx}
\usepackage{textcomp}
\usepackage{xcolor}
\usepackage{textcomp}
\usepackage{anyfontsize}
\usepackage{booktabs}
\usepackage{subfig}
\usepackage{siunitx}
\usepackage[super]{nth}
\usepackage[para]{threeparttable}
\graphicspath{{./figs/}{.}}

\newcommand{\figref}[1]{Fig.~\ref{#1}}
\newcommand{\tabref}[1]{Tbl.~\ref{#1}}
\newcommand{\secref}[1]{Sec.~\ref{#1}}

\usepackage{fancyhdr}
\fancypagestyle{mahmood}{%
  \fancyhf{} 
  
  \fancyhead[C]{\footnotesize \textcopyright 2020 IEEE. Personal use of this material is permitted.  Permission from IEEE must be obtained for all other uses, in any current or future media, including reprinting/republishing this material for advertising or promotional purposes, creating new collective works, for resale or redistribution to servers or lists, or reuse of any copyrighted component of this work in other works.}
}%
\makeatletter
\let\ps@IEEEtitlepagestyle\ps@mahmood
\makeatother

\begin{document}
    \bstctlcite{IEEEexample:BSTcontrol}

\title{
\fontsize{23}{28}\selectfont
HR-SAR-Net: A Deep Neural Network for Urban Scene Segmentation from High-Resolution SAR Data
}
    \author{\IEEEauthorblockN{
    Xiaying Wang\IEEEauthorrefmark{2}, 
    Lukas Cavigelli\IEEEauthorrefmark{2}, 
    Manuel Eggimann\IEEEauthorrefmark{2}, 
    Michele Magno\IEEEauthorrefmark{2}\IEEEauthorrefmark{3}, 
    Luca Benini\IEEEauthorrefmark{2}\IEEEauthorrefmark{3}}
    \IEEEauthorblockA{\\[-2mm]\IEEEauthorrefmark{2}ETH Zürich, Dept. EE \& IT,  Switzerland \hspace{15mm}
    \IEEEauthorrefmark{3}University of Bologna, DEI, Italy}
    \IEEEauthorblockA{\{xiaywang, cavigelli, meggimann, magnom, lbenini\}@iis.ee.ethz.ch}\vspace{-5mm}
    }
\maketitle

\begin{abstract}
Synthetic aperture radar (SAR) data is becoming increasingly available to a wide range of users through commercial service providers with resolutions reaching 0.5\,m/px. Segmenting SAR data still requires skilled personnel, limiting the potential for large-scale use. We show that it is possible to automatically and reliably perform urban scene segmentation from next-gen resolution SAR data (0.15\,m/px) using deep neural networks (DNNs), achieving a pixel accuracy of 95.19\% and a mean intersection-over-union (mIoU) of 74.67\% with data collected over a region of merely 2.2\,\si{\kilo\meter\squared}. The presented DNN is not only effective, but is very small with only 63k parameters and computationally simple enough to achieve a throughput of around 500\,Mpx/s using a single GPU. We further identify that additional SAR receive antennas and data from multiple flights massively improve the segmentation accuracy. We describe a procedure for generating a high-quality segmentation ground truth from multiple inaccurate building and road annotations, which has been crucial to achieving these segmentation results. 
\end{abstract}

    \addtolength{\textfloatsep}{-5mm}
    \addtolength{\dbltextfloatsep}{-3mm}
    \addtolength{\floatsep}{-3mm}
    \addtolength{\dblfloatsep}{-3mm}
    \addtolength{\abovedisplayskip}{-2mm}
    \addtolength{\belowdisplayskip}{-2mm}
    \linespread{0.90}
    \renewcommand{\baselinestretch}{0.90}

\section{Introduction}
Over the last few years, we have experienced a rising interest in applications of remote sensing---geospatial monitoring from space and airborne platforms. Today's sensors have seen a rapid and large improvement in spatial and spectral resolution, which has expanded the capabilities from observation of geological, atmospheric, and vegetation phenomena to applications such as the extraction of high-resolution elevation models and maritime vessel tracking \cite{Huang2018,Moumtzidou2019,Chouhan2018,Xia2017,Chen2019a}.

One specific remote sensing method, synthetic aperture radar (SAR) imaging, uses a moving, side-looking radar which transmits electromagnetic pulses and sequentially receives the back-scattered signal \cite{Zelnio2018}. This time series of amplitude and phase values contains information on the relative location, surface geometry/roughness, and permittivity of the reflecting objects. 
The frequency of the transmitted electromagnetic signal defines the penetration depth into the soil, and the bandwidth tunes the geometric resolution. 
SAR is particularly interesting for a wide range of applications as it is---as opposed to visual or multispectral imaging methods---resilient to weather conditions and cloud coverage, and it is independent of the lighting conditions (night, dusk/dawn) \cite{Wang2018}. By a mere overflight of such a sensor either aboard a satellite, airplane, or drone, data can be collected to monitor crop growth, ocean wave height, floating icebergs, biomass estimation, snow monitoring, and maritime vessel tracking. 

This wide range of applications and the high reliability have lead to a rapid increase in SAR imaging satellites orbiting the Earth. It has fueled the rise of newly-founded data providers such as ICEYE and Capella Space with the intention of providing such imaging data for commercial purposes at resolutions of up to 0.5\,m/px and with a schedule to reach hourly re-visitation rates of any point globally within the next few years. 

Given the growing amount of SAR data, it is critical to develop automated analysis methods, possibly in real-time pipelines. Deep neural networks (DNNs) play a key role in this automation effort. They have become the method of choice for many computer vision applications over the last few years, pushing the accuracy far beyond previous methods and exceeding human accuracy on tasks such as image recognition. DNNs have also shown state-of-the-art performance in image segmentation in application scenarios from road segmentation for autonomous driving to tumor segmentation in medical image analysis and road network extraction from visual imagery. 

The main contributions of this work are three-fold: 1) presenting a dataset with high-resolution (0.15\,m/px) SAR imaging data and ground truth annotations for urban land-use segmentation, 2) proposing and evaluating deep neural network topologies for automatic segmentation of such data, and 3) providing a detailed analysis of the segmentation results and which input data/features are most beneficial for the quality of results, reducing the error rate from 16.0\% to 4.8\% on relative to related work for a similar task and reaching a mean IoU of 74.67\%.

\section{Related Work}
In this section, we first introduce existing SAR datasets and then provide an overview of current analysis methods. 

\subsection{SAR Datasets}
Probably the best known SAR dataset for classification and detection purposes is the MSTAR dataset of the U.S. air force from 1995/1996 for target recognition based on 30\,cm/px X-band images with HH polarization. It contains images with $54\times54$ to $192\times192$\,px of 15 types of armored vehicles with a few hundred images each and a few large scenes with which more complex scenes can be composed for detection tasks. Nevertheless, such detection tasks are not very hard with clearly distinct objects in an open field and often simplified classes (e.g., personnel carrier/tank/rocket launcher/air defense unit). State-of-the-art methods achieve a 100\% detection and recognition rate on such generated detection problems \cite{Yang2019a,Zelnio2018}. 

Another large SAR dataset is OpenSARShip 2.0 \cite{Huang2018}. It uses 41 images of the ESA's Sentinel-1 mission and includes labels of $\approx11000$ ships of various types, of which 8470 are cargo ships, and 1740 are tankers. It uses C-band SAR to generate 10\,m/px resolution images. 

Several recent SAR datasets focus on providing matching pairs of SAR and visual images: the TUM/DLR SEN1-2 dataset from 2018 \cite{Schmitt2018} provides 10\,m/px data from Sentinel-1 and 2, and the SARptical dataset (also 2018) \cite{Wang2018e} provides TerraSAR-X data at a 1\,m/px resolution including extracted 3D point clouds and matching visual images. 

Other datasets such as \cite{Abdikan2016} or JAXA's ALOS PALSAR forest/non-forest \cite{Shimada2014} (18\,m/px) typically come with a much lower resolution (\textgreater10\,m/px) and often for applications in large-scale land cover analysis with classes such as water/forest/urban/agriculture/bareland. Most SAR datasets, particularly more recent ones with improved resolution, do not come with any ground truth annotations. Such datasets include the SLC datasets of ICEYE with 1\,m/px resolution X-band SAR images with a VV polarization, publicly available images of the DLR/Airbus TerraSAR-X satellite, etc. Such datasets must first be combined and registered to, or annotated with ground truth labels.

\subsection{SAR Data Analysis Methods}
Traditional signal processing methods still dominate the field of SAR data analysis. While in some cases, properties specific for SAR data and corresponding particular sensor configurations are used, e.g., for change detection \cite{MendezDominguez2018,MendezDominguez2019}. Also, tasks focusing on semantic segmentation are widely addressed using statistical models and engineered features \cite{Belloni2017,Duan2018,Tschannen2016}. 

DNNs have achieved excellent results segmenting visual data, including satellite images, for geostatistical uses as well as disaster-relief (e.g., road passability estimation) \cite{Moumtzidou2019,Chouhan2018,Xia2017,Chen2019a}. 
To leverage these results, several researchers have created datasets and proposed methods for SAR-to-visual translation and SAR data generation from visual images using generative adversarial networks (GANs) or using DNNs on visual data to generate labels for training DNNs on SAR data \cite{Reyes2019,Wang2018d,Liu2018c}. 
These works provide helpful methods to rapidly obtain labels for SAR datasets or enlarge such datasets with generated samples at a reduced cost. However, the resulting data is inherently less accurate than systematically collected ground truth labels. 

Several efforts have also been undertaken towards urban land-use segmentation. 
In \cite{Yao2017}, they compare the building segmentation accuracy of an atrous ResNet-50 DNN using OSM ground truth and training the same network on TerraSAR-X data with a 2.9\,m/px resolution and visual data extracted from Google Earth, achieving a pixel-wise accuracy of 74\% and 82.9\%, respectively. 
In \cite{Shahzad2019}, the authors propose an FCN-style \cite{Long2015} network for building segmentation and have achieved a pixel accuracy of 78\%--92\%. 
In \cite{Henry2018}, they segment roads using the FCN-8s and DeepLabv3+ networks on 1.25\,m/px TerraSAR-X images, achieving pixel accuracies of 43\%--46\%. They have determined OSM annotations being too inaccurate and manually labeled all the roads based on human analysis of the SAR data. Further, they introduced \emph{spatial tolerance} for the road classification, not considering a stripe of 1--8 around the boundary between road and non-road segments. 

The most similar work to ours is \cite{Wu2019a}. They use DNNs to segment PolSAR data from \cite{Wu2018b} into \emph{built-up area}, \emph{vegetation}, \emph{water}, \emph{road}, and \emph{others}. They achieve a pixel accuracy of 84\% and 84\% and a mean accuracy of 64\% and 51\% using an FCN \cite{Long2015} and U-net \cite{Ronneberger2015} type of DNN, respectively.

\section{Dataset Creation}

\begin{figure*}[!t]
    \def\imgTileWidth{0.248\linewidth}
    \centering
    \subfloat[]{\includegraphics[trim={0cm 0cm 0cm 0cm}, clip=true, width=\imgTileWidth]{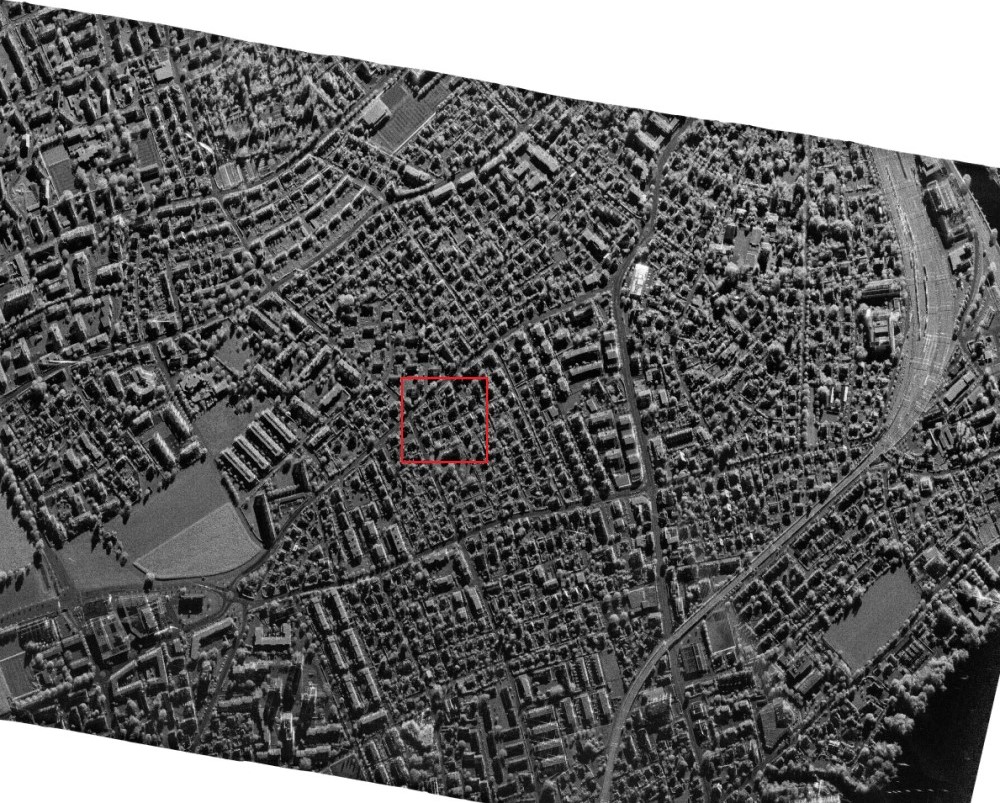}\label{fig:sarImg}}
    \hfill
    \subfloat[]{\includegraphics[trim={2cm 1cm 2cm 1cm}, clip=true,  width=\imgTileWidth]{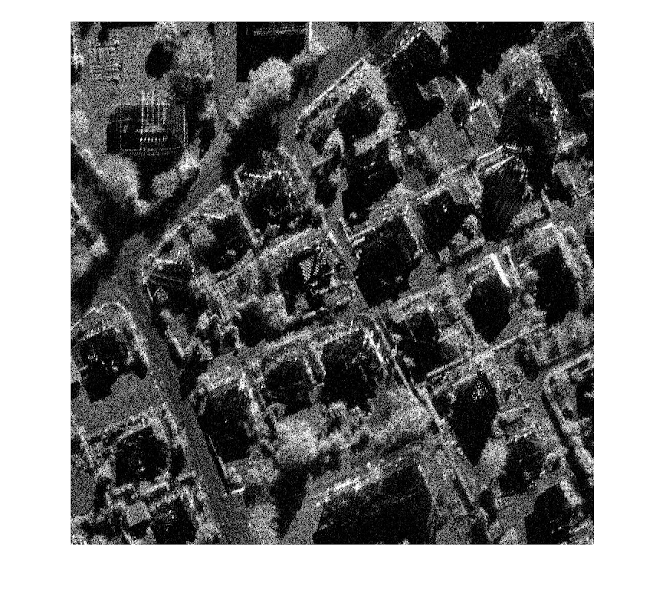}\label{fig:sarImgZoom}} 
    \hfill
    \subfloat[]{\includegraphics[trim={2cm 1cm 2cm 1cm}, clip=true, width=\imgTileWidth]{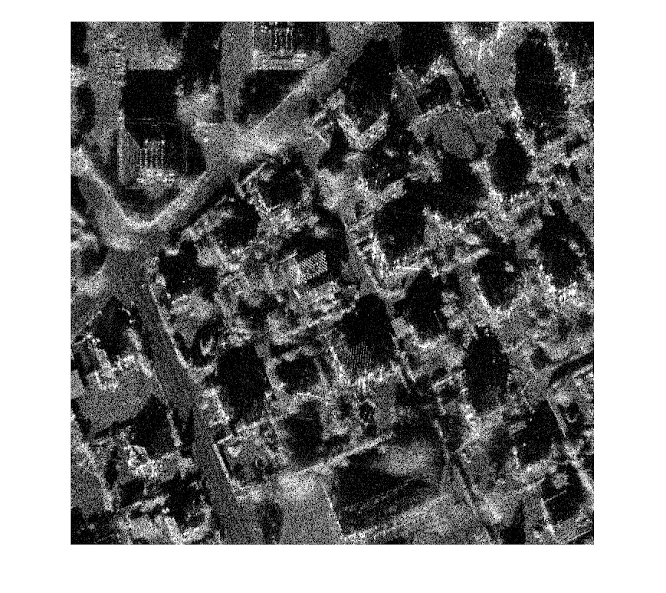}\label{fig:sarImg004Zoom}}
    \hfill
    \subfloat[]{\includegraphics[clip=true, width=\imgTileWidth]{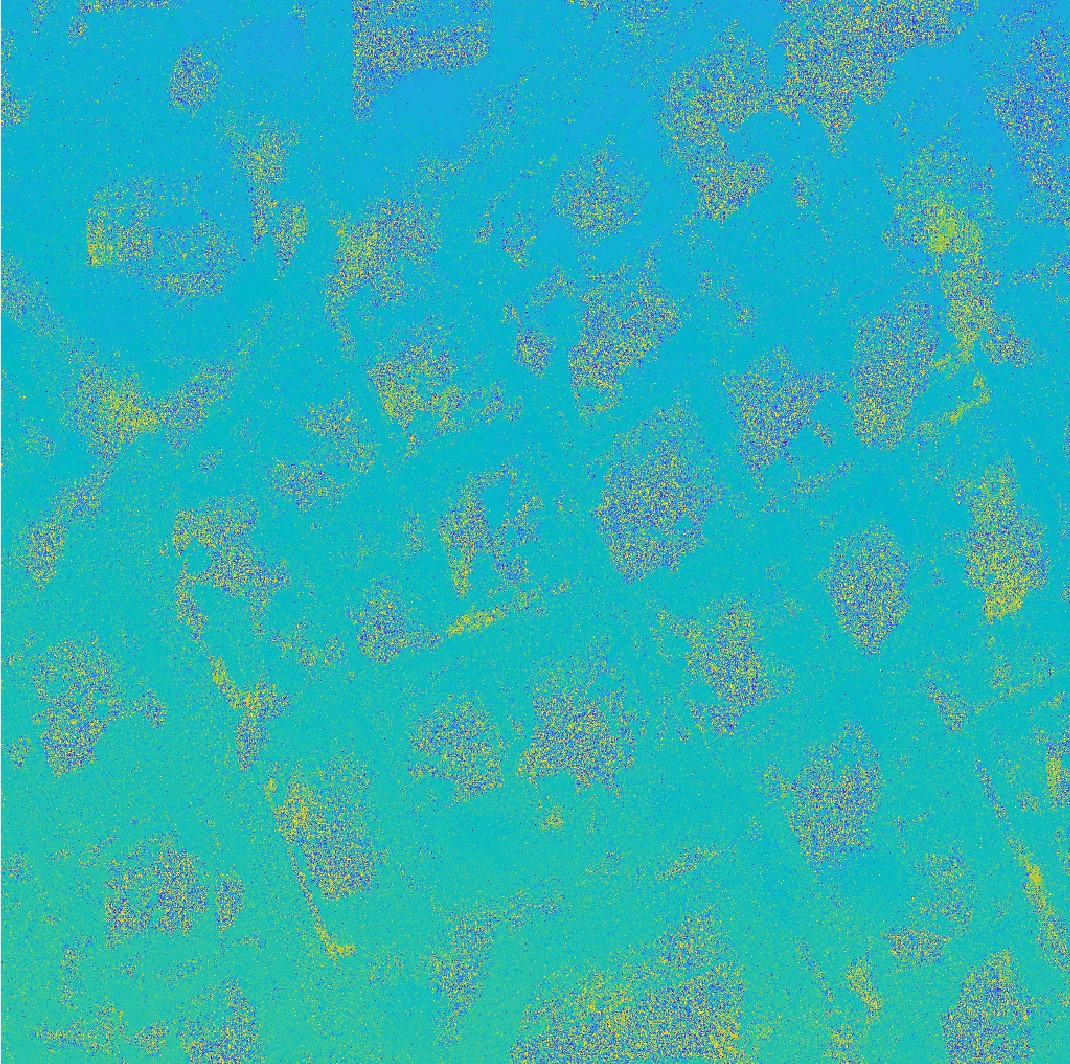}\label{fig:phaseDiffCos}}
    \vspace{-2mm}
    \subfloat[]{\includegraphics[trim={2cm 1cm 2cm 1cm}, clip=true, width=\imgTileWidth]{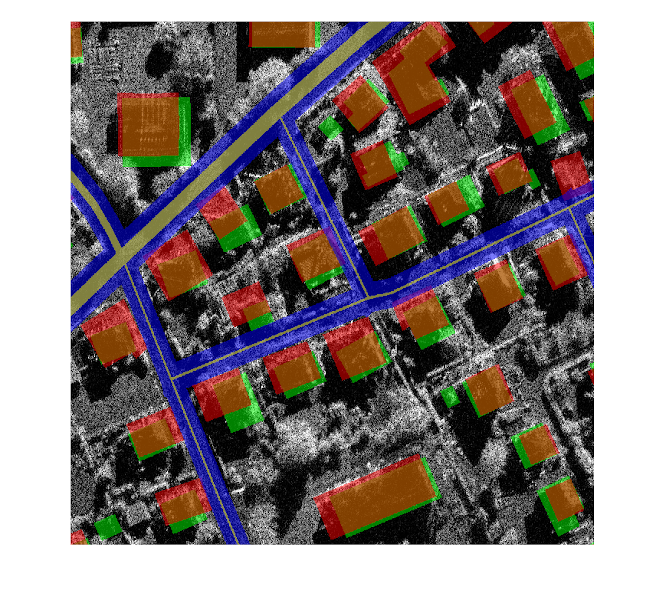}\label{fig:exampleImgLabel}}
    \hfill
    \subfloat[]{\includegraphics[trim={2cm 1cm 2cm 1cm}, clip=true, width=\imgTileWidth]{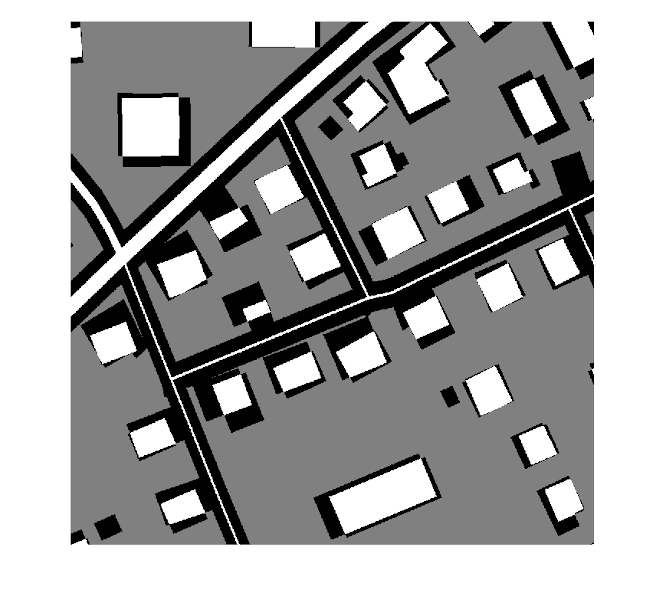}\label{fig:exampleLabel}}
    \hfill
    \subfloat[]{\includegraphics[trim={2cm 1cm 2cm 1cm}, clip=true, width=\imgTileWidth]{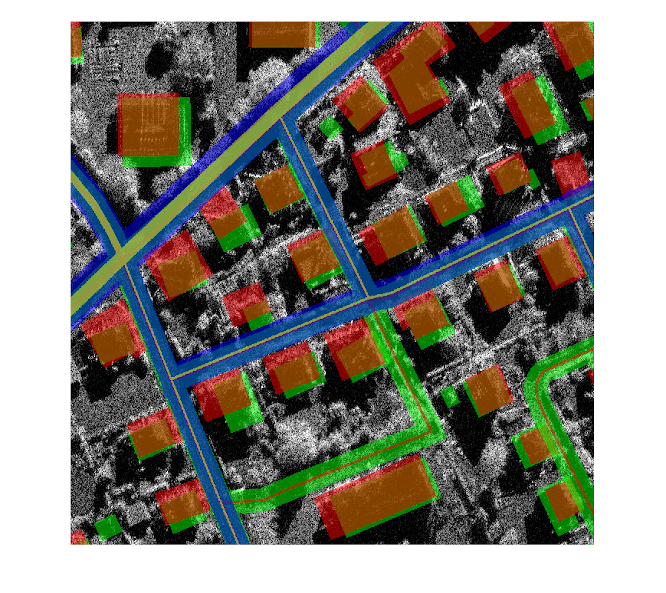}\label{fig:exampleImgLabelOsmGovtRoads}}
    \hfill
    \subfloat[]{\includegraphics[trim={2cm 1cm 2cm 1cm}, clip=true, width=\imgTileWidth]{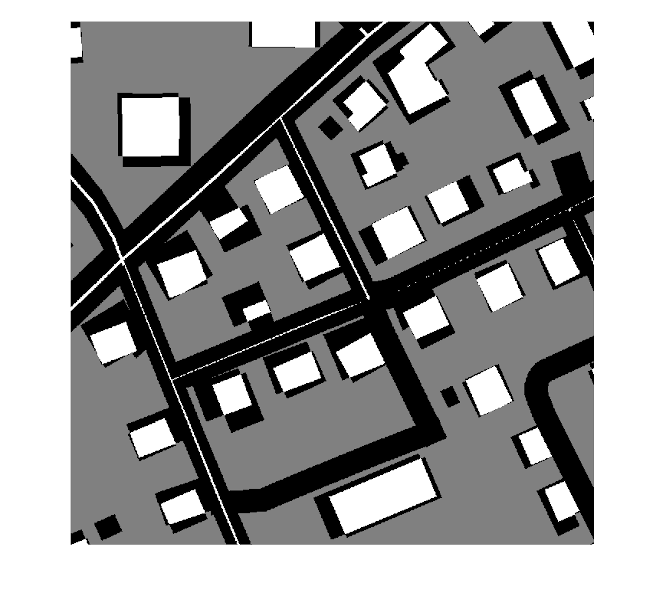}\label{fig:exampleLabelOsmGovtRoads}}
    \caption{(a) Overview of an entire SAR recording. (b) Close-up view of 1024$\times$1024 of SAR data. The region shown is highlighted in the overview. (c) Same location from the second SAR image taken at a different angulation. (d) Cosine of phase difference between channels 1 and 4. (e) Overlapped ground truth labels (green: OSM buildings, red: swissTLM3D buildings, blue/yellow: wide/narrow OSM roads). (f) Ground truth (white: buildings/roads, gray: other, black: unclassified). (g) Added minor roads from both OSM (blue/yellow) and swissTLM3D map (green/red). (h) Corresponding ground truth image.}
    \label{fig:exampleImages}
\end{figure*}

\subsection{SAR Data} \label{sec:sarDataRecording}
The data was recorded in Autumn 2017 by \emph{armasuisse} and the University of Zurich's SARLab during two flights over Thun, Switzerland, with an off-nadir angle of 55\textdegree{} and with a 35\,GHz carrier (Ka-band) using the Fraunhofer/IGI Miranda-35 sensor. From the recorded data, a rasterized ``image'' with a $0.15\times0.15$\,m/ps resolution was generated, including phase information. The system contains 4 channels, each with data recorded from a separate receiving antenna. 
We do not apply any preprocessing methods to compensate foreshortening or layover, or to reduce the speckle noise \cite{Moreira2013}. 
The resulting images each cover an area of 2.2\,\si{\square\kilo\meter} with 97.17\,Mpx and have been recorded during two flights on parallel trajectories on opposite sides of the region-of-interest with the resulting images overlapping. The second recording provides data for areas not visible to the radar due to occlusion (cf. \figref{fig:sarImgZoom} and \figref{fig:sarImg004Zoom}). 

\subsection{Labeling} \label{sec:datasetLabeling}
High-quality ground truth data is crucial to train effective DNNs: it not only severely affects the quality of the results, but also has a substantial impact on training time and generalization of the trained model \cite{Zlateski2018}. For this work, we consider three classes---building, road, other---and allow for some pixel to be annotated as unlabeled. 

We have used OSM data on buildings and roads to generate the ground truth segmentation. In order to further improve the quality, we fused this information with the building shape annotations of the government-created \emph{swisstopo swissTLM3D} map. \figref{fig:exampleImgLabel} visualizes the annotations of both data sources on a tile of the dataset. The annotations mostly overlap, but significant differences exist with sometimes one of the sources missing a building entirely. Visual inspection showed that none of these sources reliably captures all buildings. We have thus merged the annotations, to only classify a pixel as \emph{building} or not-a-building where the sources agree, and left the pixels with unclear annotations as unlabeled (cf. \figref{fig:exampleImgLabel}). 

The roads captured in OSM, as well as the \emph{swisstopo} data, are provided as line segments with several rank annotation (for OSM: motorway, path, pedestrian, platform, primary, residential, secondary, service, steps, tertiary, track, bus stop, cycleway, footway, living street, unclassified). While this information aids in estimating the width of the road required to create the segmentation ground truth, a very wide variance remains within each category. We thus assign two widths to each type of road: an expected maximum width and a minimum width. The latter is used to annotate pixels as \emph{road}, whereas the former is used to mark the surrounding pixels as unlabeled, similar to \cite{Henry2018}. 

With roads of this many different ranks and sizes, some of them are of minor relevance and hard to recognize due to the small size. We thus create three different annotations, which we then evaluate experimentally in \secref{sec:results}:
\begin{enumerate}
    \item Annotating only the main roads from OSM: primary, secondary, tertiary, residential, and living street (cf. \figref{fig:exampleImgLabel} and \figref{fig:exampleLabel}), 
    \item Annotating all the roads from OSM including the minor ones, and
    \item Combining the annotations from OSM and swissTLM3D using the same rule for fusing the data as for the buildings: labeling roads only where both sources agree (cf. \figref{fig:exampleImgLabelOsmGovtRoads} and \figref{fig:exampleLabelOsmGovtRoads}). We can see that some roads, especially the minor ones, have a significant offset between the two map sources, leaving some of them entirely unlabeled.
\end{enumerate}

\section{Algorithm}

\subsection{DNN Topology}
\begin{figure*}
    \centering
    \includegraphics[trim={0cm 4.3cm 0cm 3.5cm}, clip=true, width=0.95\linewidth]{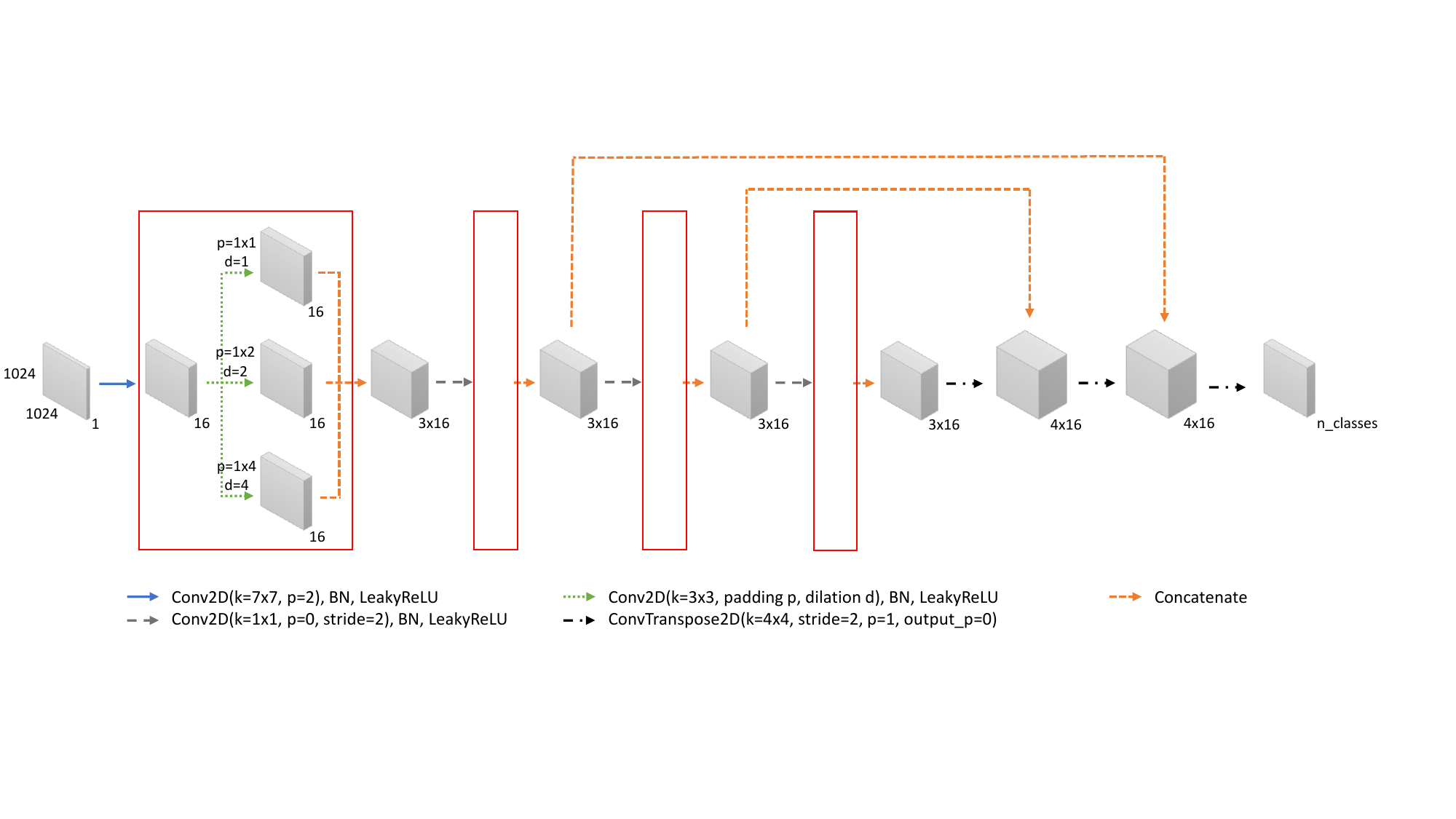}
    \caption{The architecture of HR-SAR-Net.}
    \label{fig:dnnArch}
\end{figure*}
We show the proposed DNN topology in \figref{fig:dnnArch}.  After a first 2d convolution from one or more input feature maps (depending on the selected features) to 16, the feature maps are passed through one of our basic building blocks: 3 parallel $3\times3$ convolutions with 16 output feature maps and a dilation factor 1, 2 and 4, respectively, of which the results are concatenated, followed by a $1\times1$ convolution to reduce the number of feature maps back to 16. This allows capturing information from various context sizes in each such block. This is followed by the common U-Net structure introduced in \cite{Ronneberger2015}: the feature maps are fed through several convolution layers, some of which are strided, followed by several transposed convolution layers to recover the resolution, after each of which the last of the previous sets of feature maps is concatenated to recover the high-resolution information in order to allow fine-grained segmentation. 

The presented network is light-weight in terms of parameter count and computation effort compared to other segmentation networks. Due to the low number of feature maps throughout the network, the small filter sizes, and the use of dilated convolutions and striding to cover the required field of view, it has merely 63k parameters. It requires 13k multiply-accumulate (MAC) operations per pixel for inference\footnote{Note that the stride of the first $1\times1$ convolution layer can be brought forward to the dilated convolution layer with identical results.}, which means for a relatively large input of $1024\times1024$\,px it requires 13.6G MAC operations. For comparison, recent GPUs such as Nvidia's GTX 2080 Ti can perform 6.7T single-precision MAC operations per second and can thus in a very rough approximation process 515\,Mpx/s. 

\subsection{Feature Extraction}
Our overall SAR data is comprised of two recordings (cf. \secref{sec:sarDataRecording}), where each has multiple channels (receive antennas), as well as magnitude and phase information. We extract the following features: 
\begin{enumerate}
    \item Magnitude information: The main information, and a visually human-interpretable image, is obtained by 1) compressing the magnitude values of any of the SAR images of all the recordings and channels to log-scale, 2) clamping the maximum value at the \nth{99} percentile, and 3) limiting the range to 25\,dB. An image after such preprocessing is shown in \figref{fig:sarImgZoom}. 
    \item Phase information: Each pixel also records the phase of the reflected signal. Such data can be used to obtain further information on the relative distance. In order to convert it to a real-valued feature map, we either used the real and imaginary part of the length-normalized complex vector of each pixel (named cos and sin phase hereafter), or the real and imaginary parts of the dB-scaled magnitude rotated by the phase of the original pixel value (named re/im hereafter).
    \item Phase difference information: As opposed to the other features, which can be applied to each recording and channel individually, we here take the phase difference of each pixel between two channels (channels 1 and 4, unless specified otherwise). Visually, such a feature map is indicative of radar shadows and structure height. We convert this data into two real-valued feature maps identically to the phase information. The cosine/real component of the phase difference is shown in \figref{fig:phaseDiffCos}.
\end{enumerate}

By combining the magnitude feature map (FM) with the 4 phase FMs for each channel and both recordings, plus the two phase difference FMs for each recording, we obtain a total of 44 FMs. However, these features are redundant, require a sizable amount of memory, and might only lead to overfitting. We thus perform feature selection based on the experimental results in \secref{sec:results}. 

\section{Results \& Discussion} \label{sec:results}

\begin{table*}
\caption{Classification Accuracy and Feature Selection}
\label{tbl:resultsOverview}
\centering
\begin{threeparttable}
\setlength{\tabcolsep}{3pt}
\begin{tabular}{lccccccccccccccccc}
\toprule
experiment no. & 1 & 2 & 3 & 4 & 5 & 6 & 7 & 8 & 9 & 10 & 11 & \textbf{12} & 13 & 14\tnote{*} & 15\tnote{*} & \textbf{16}\tnote{$\dagger$} & 17\tnote{$\dagger$} \\ 
\midrule
flights & 1 & 1 & 1 & 1 & 1 & 1 & 1 & 1 & 1,2 & 1,2 & 1,2 & 1,2 & 1,2 & 1,2 & 1,2 & 1,2 & 1,2 \\
channels & 1 & 1 & 1 & 1 & 1--4 & 1--4 & 1--4 & 1--4 & 1 & 1 & 1 & 1--4 & 1--4 & 1--4 & 1--4 & 1--4 & 1--4 \\
magnitude & ✓ & ✓ &  & ✓ & ✓ & ✓ &  & ✓ & ✓ & ✓ &  & ✓ &  & ✓ &  & ✓ &  \\
phase: cos/sin &  & ✓ &  &  &  & ✓ &  &  &  & ✓ &  &  &  &  &  &  &  \\
phase: re/im &  &  & ✓ &  &  &  & ✓ &  &  &  & ✓ &  & ✓ &  & ✓ &  & ✓ \\
phase diff. &  &  &  & ✓ &  &  &  & ✓ &  &  &  &  &  &  &  &  &  \\
\midrule
pixel accuracy [\%] & 89.26 & 85.97 & 83.80 & 85.78 & 89.37 & 85.22 & 90.50 & 86.85 & 90.03 & 84.89 & 86.64 & \textbf{91.89} & 91.78 & 90.81 & 90.11 & \textbf{95.19} & 93.49 \\
mean accuracy [\%] & 76.31 & 66.52 & 63.76 & 66.14 & 72.89 & 65.00 & 80.78 & 68.22 & 71.43 & 65.45 & 67.65 & \textbf{75.12} & 79.00 & 72.21 & 72.29 & \textbf{90.30} & 86.12 \\
mean IoU [\%] & 56.92 & 59.02 & 55.33 & 57.03 & 64.15 & 55.48 & 68.43 & 59.34 & 65.79 & 58.09 & 60.21 & \textbf{70.18} & 69.69 & 66.35 & 66.05 & \textbf{74.67} & 69.90 \\
\bottomrule
\end{tabular}
\begin{tablenotes}\footnotesize
\item[*] Including \emph{swisstopo} annotations for roads. 
\item[$\dagger$] Without class balancing.
\end{tablenotes}
\end{threeparttable}
\end{table*}

\begin{figure}
    \centering
    \subfloat[]{\includegraphics[trim={2cm 2cm 2cm 2cm}, clip=true, width=0.5\linewidth]{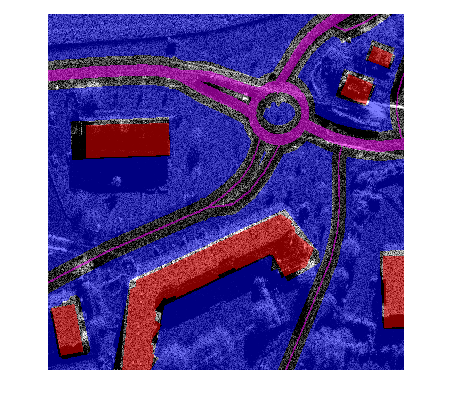}\label{fig:43truth16}}
    \hfill
    \subfloat[]{\includegraphics[trim={2cm 2cm 2cm 2cm}, clip=true,  width=0.5\linewidth]{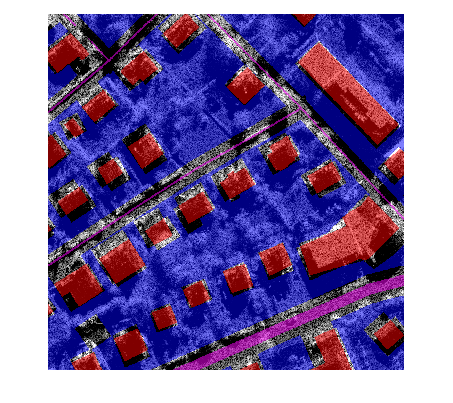}\label{fig:43truth22}} 
    \vfill
    \vspace{-2mm}
    \subfloat[]{\includegraphics[trim={2cm 2cm 2cm 2cm}, clip=true, width=0.5\linewidth]{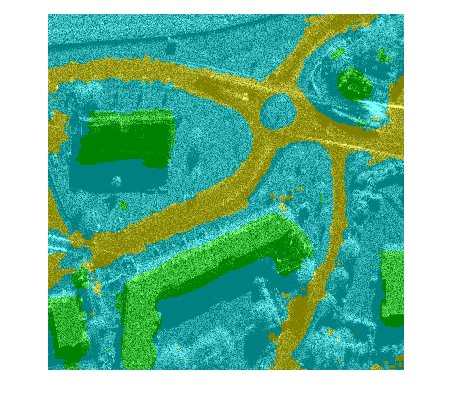}\label{fig:43pred16}} 
    \hfill
    \subfloat[]{\includegraphics[trim={2cm 2cm 2cm 2cm}, clip=true, width=0.5\linewidth]{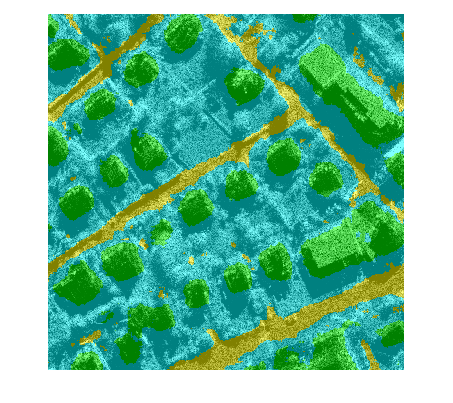}\label{fig:43pred22}}
    \vfill
    \vspace{-2mm}
    \subfloat[]{\includegraphics[trim={2cm 2cm 2cm 2cm}, clip=true, width=0.5\linewidth]{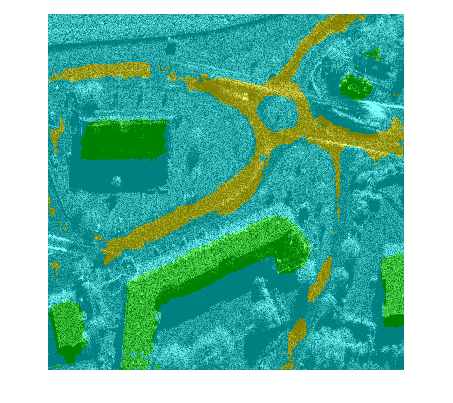}\label{fig:55pred16}} 
    \hfill
    \subfloat[]{\includegraphics[trim={2cm 2cm 2cm 2cm}, clip=true, width=0.5\linewidth]{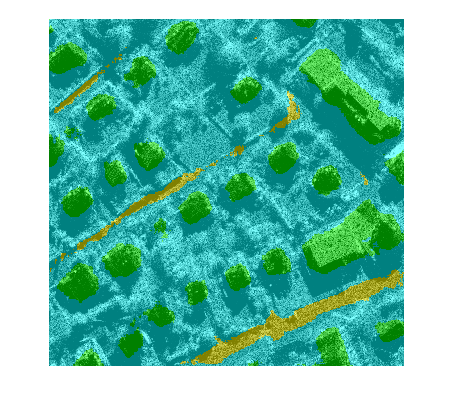}\label{fig:55pred22}}
    \caption{Qualitative segmentation results on the test set. Ground truth (a,b), predictions from experiments 12 (c,d) and 16 (e,f).}
    \label{fig:outputAndLabels}
\end{figure}


\subsection{Experimental Setup}
For our experiments, we have split the SAR images into spatial tiles of $1024\times1024$. The resulting 86 tiles are split into a training set of 64 tiles (74\%) and a test set of 22 tiles (26\%). The networks were trained with PyTorch using the Adam optimizer. Starting from an initial learning rate of $10^{-2}$, we used a reduce-on-plateau learning rate schedule (factor: 0.1, patience: 10, relative threshold: $10^{-4}$). The batch size was 8 and split across 4 GTX 2080 Ti GPUs. We use the cross-entropy loss function for all experiments. Unless otherwise noted, we have weighted the classes in the loss function to compensate for the inherent class imbalance. The experiments have, in general, fully converged after a maximum of 80 epochs, corresponding to an average training time of merely 8 minutes. Following related work, we use the metrics \emph{pixel accuracy (PA)}, \emph{mean accuracy (MA)}, and \emph{mean intersection-over-union (mIoU)}, 
\begin{align}
    \mathrm{PA} &= \frac{\sum_i n_{i,i}}{\sum_i \sum_j n_{i,j}}, \quad
    \mathrm{MA} = \frac{1}{C}\sum_i \frac{n_{i,i}}{\sum_j n_{i,j}}, \\
    \mathrm{mIoU} &= \frac{1}{C}\sum_i \frac{n_{i,i}}{\sum_j n_{i,j} + \sum_j n_{j,i} - n_{i,i}}, 
\end{align}
where $n_{i,j}$ is the number of pixels with target class $i$ and predicted class $j$, and $C$ is the number of classes. The mIoU is a widely used metric for segmentation, since it adds weight to classes covering only few pixels and yet caputures the false positives and false negatives of each class.

\subsection{Feature Selection}\label{sec:resultsFeatureSel}
We have trained and evaluated the network for many combinations of features. An overview of the results is shown in \tabref{tbl:resultsOverview}. For the phase information features, we can observe that in each of the experiment groups 1--4, 5--8, 9--11, and 12--13, providing phase difference information or phase information in the cos/sin representation consistently resulted in worse results than using only the magnitude features. Using only the re/im features without the magnitude feature has shown similar performance to providing the magnitude feature map, although generally slightly worse with the exception of the exp. group 5--8. For the remaining analyses, we thus use the magnitude features solely. 

As for using multiple channels, we see a clear accuracy gain comparing experiments 1 to 5 and 9 to 12 of 1.73\% and 1.86\%, respectively. Similarly, for using data from both flights, we observe improvements by 2.39\% and 2.52\% from experiments 1 to 9 and 5 to 12. 

\subsection{Ground Truth Selection}
In \secref{sec:datasetLabeling}, we have left the decision on which annotations to use for the roads to experimental evaluation. Exp. 14 and 15 include the \emph{swisstopo} road annotations in addition to the OSM data used for the other experiments. However, they do not further improve the accuracy but instead, reduce it. Each additional road comes with several pixels marked as unlabeled in its surroundings, and roads are only labeled as such if the ground truths agree, This might leave the impression of simplifying the classification task. However, requiring the agreement of both annotations sources also implies removing some valuable labeled \emph{road} pixels from the training data. We thus attribute this small accuracy drop to the latter effect outweighing the slight simplification of the task and proceed using only OSM annotations to label the roads. 

\begin{table*}
    \centering
    \caption{Comparison to Related Work}
    \label{tab:compRelWork}
    \begin{tabular}{lccccc}
        \toprule
           & \cite{Yao2017} & \cite{Yao2017} & \cite{Wu2019a} & \cite{Henry2018} & ours \\ 
        \midrule
          sensor/source & TerraSAR-X & Google Earth & PolSAR & TerraSAR-X & Miranda-35 \\ 
          resolution & 2.9\,m/px & 2.9\,m/px & 0.5\,m/px & 1.25\,m/px & 0.15\,m/px \\
          labels & OSM & OSM & human & human & OSM \& swisstopo \\ 
          classes & bldg/landuse/natural/water & bldg/landuse/natural/water & road/water/bldg/vegetation & road/other & bldg/road/other \\ 
          DNN type & Atrous-ResNet50 & Atrous-ResNet50 & FCN & FCN & mod. U-Net \\ \midrule
          pixel accuracy & 74.0\% & 82.9\% & 84.0\% & --- & 95.2\% \\
          mean accuracy & --- & --- & 64.0\% & --- & 90.3\% \\
          mean IoU & 30.0\% & 43.7\% & 50.0\% & 45.6\% & 74.7\% \\
          qualitative acc. & bad & bad & ok & very good & very good \\
        \bottomrule
    \end{tabular}
\end{table*}

\subsection{Overall Accuracy and Class Balancing}
To compensate potentially averse effects on the results due to imbalanced classes, we assess using a class-balanced loss function during training by weighting pixels inversely to their rate of occurence. 
Following our insights on which features to use from \secref{sec:resultsFeatureSel}, we use data from all channels and both flights but no phase information, to achieve a pixel accuracy of 91.89\% and a mIoU of 70.18\%. If we do not compensate for the class imbalance, the pixel accuracy and the mIoU further increase to 95.19\% and 74.67\%. For a more qualitative analysis, we provide some example results from the best predictor with and without class balancing alongside the ground truth information in \figref{fig:outputAndLabels}. Particularly for the network trained with class balancing, we can see an outstanding segmentation quality. The buildings are segmented very well, and most misclassified pixels are observed at the segments classified as road leaking into the driveways or marking very small clusters of pixels in the backyards as road. For the class-balanced network, the \emph{road} class is assigned much less frequently, as a misclassification as \emph{other} is much more penalized, and the pixels labeled as road are much fewer---particularly on smaller roads. As the class \emph{road} is not as strictly defined as \emph{building}---private roads, forecourts, and driveways are labeled as \emph{other} although they are not distinguishable from their public pendants to a human annotator either---the decision of the resulting classifier can be expected and observed to be less confident as well.

\subsection{Comparison to Related Work}
We compare our results to related work in \tabref{tab:compRelWork}. We see a vast improvement in accuracy, reducing the error rate from 16\% to 4.8\% on a similar task. Akin, we observe gains in mean accuracy and mIoU. A qualitative comparison of the segmentation outputs shows vast improvements (cf. \figref{fig:outputAndLabels} and \cite{Yao2017,Henry2018,Wu2019a}). The main differences to the other methods is the fusion of two annotation sources, the vastly increased resolution, the two 4-channel recordings from opposite directions, as well as the optimized DNN. 

\section{Conclusion}
We have proposed and evaluated a DNN to automatically and reliably perform urban scene segmentation from high-resolution SAR data, achieving a pixel accuracy of 95.2\% and a mean IoU of 74.7\% with data collected over a region of merely 2.2\,\si{\kilo\meter\squared}. The presented DNN is not only effective, but is very small with only 63k parameters and computationally simple enough to achieve a throughput of around 500\,Mpx/s using a single GPU. We have further identified that additional SAR receive antennas and data from multiple flights massively improve the segmentation accuracy while phase information showed no positive effect. The procedure described for generating a high-quality segmentation ground truth from multiple inaccurate building and road annotations has shown to be crucial to achieve good segmentation results.

    \section{Acknowledgements}
    We would like to thank \emph{armasuisse Science and Technology} for funding this research and for providing---jointly with the University of Zurich's SARLab---the SAR data used throughout this work. 

\bibliographystyle{IEEEtran}
\bibliography{myBstCtl, bib-lukas}

\begin{thebibliography}{10}
\providecommand{\url}[1]{#1}
\csname url@samestyle\endcsname
\providecommand{\newblock}{\relax}
\providecommand{\bibinfo}[2]{#2}
\providecommand{\BIBentrySTDinterwordspacing}{\spaceskip=0pt\relax}
\providecommand{\BIBentryALTinterwordstretchfactor}{4}
\providecommand{\BIBentryALTinterwordspacing}{\spaceskip=\fontdimen2\font plus
\BIBentryALTinterwordstretchfactor\fontdimen3\font minus
  \fontdimen4\font\relax}
\providecommand{\BIBforeignlanguage}[2]{{%
\expandafter\ifx\csname l@#1\endcsname\relax
\typeout{** WARNING: IEEEtran.bst: No hyphenation pattern has been}%
\typeout{** loaded for the language `#1'. Using the pattern for}%
\typeout{** the default language instead.}%
\else
\language=\csname l@#1\endcsname
\fi
#2}}
\providecommand{\BIBdecl}{\relax}
\BIBdecl

\bibitem{Huang2018}
L.~Huang \emph{et~al.}, ``{OpenSARShip: A Dataset Dedicated to Sentinel-1 Ship
  Interpretation},'' \emph{IEEE Journal of Selected Topics in Applied Earth
  Observations and Remote Sensing}, vol.~11, no.~1, pp. 195--208, 2018.

\bibitem{Moumtzidou2019}
A.~Moumtzidou \emph{et~al.}, ``{Road Passability Estimation Using Deep Neural
  Networks and Satellite Image Patches},'' in \emph{Proc. BiDS}.\hskip 1em plus
  0.5em minus 0.4em\relax European Commission, 2019.

\bibitem{Chouhan2018}
S.~S. Chouhan, A.~Kaul, and U.~P. Singh, ``{Soft computing approaches for image
  segmentation: a survey},'' \emph{Multimedia Tools and Applications}, vol.~77,
  no.~21, pp. 28\,483--28\,537, 2018.

\bibitem{Xia2017}
X.~Xia and B.~Kulis, ``{W-Net: A Deep Model for Fully Unsupervised Image
  Segmentation},'' \emph{arXiv:1711.08506}, 2017.

\bibitem{Chen2019a}
G.~Chen \emph{et~al.}, ``{Fully Convolutional Neural Network with Augmented
  Atrous Spatial Pyramid Pool and Fully Connected Fusion Path for High
  Resolution Remote Sensing Image Segmentation},'' \emph{Applied Sciences},
  vol.~9, no.~9, p. 1816, 2019.

\bibitem{Zelnio2018}
E.~G. Zelnio, M.~Levy, R.~D. Friedlander, and E.~Sudkamp, ``{Deep learning
  model-based algorithm for SAR ATR},'' in \emph{Proc. SPIE Algorithms for
  Synthetic Aperture Radar Imagery XXV}, 2018.

\bibitem{Wang2018}
X.~Wang \emph{et~al.}, ``{Embedded Classification of Local Field Potentials
  Recorded from Rat Barrel Cortex with Implanted Multi-Electrode Array},'' in
  \emph{Proc. IEEE BIOCAS}, 2018.

\bibitem{Yang2019a}
Z.~Cui, C.~Tang, Z.~Cao, and N.~Liu, ``{D-ATR for SAR Images Based on Deep
  Neural Networks},'' \emph{Remote Sensing}, vol.~11, no.~8, 2019.

\bibitem{Schmitt2018}
M.~Schmitt, L.~H. Hughes, and X.~X. Zhu, ``{The SEN1-2 Dataset for Deep
  Learning in SAR-Optical Data Fusion},'' \emph{ISPRS Annals of the
  Photogrammetry, Remote Sensing and Spatial Information Sciences}, vol.~4,
  no.~1, pp. 141--146, 2018.

\bibitem{Wang2018e}
Y.~Wang and X.~X. Zhu, ``{The SARptical Dataset for Joint Analysis of SAR and
  Optical Image in Dense Urban Area},'' in \emph{Proc. IEEE IGARSS}, 2018, pp.
  6840--6843.

\bibitem{Abdikan2016}
S.~Abdikan, F.~B. Sanli, M.~Ustuner, and F.~Cal{\`{o}}, ``{Land Cover Mapping
  Using Sentinel-1 SAR Data},'' \emph{ISPRS - International Archives of the
  Photogrammetry, Remote Sensing and Spatial Information Sciences}, vol.
  XLI-B7, no. July, pp. 757--761, 2016.

\bibitem{Shimada2014}
M.~Shimada \emph{et~al.}, ``{New global forest/non-forest maps from ALOS PALSAR
  data (2007–2010)},'' \emph{Remote Sensing of Environment}, vol. 155, pp.
  13--31, 2014.

\bibitem{MendezDominguez2018}
E.~Mendez~Dominguez \emph{et~al.}, ``{A Multisquint Framework for Change
  Detection in High-Resolution Multitemporal SAR Images},'' \emph{IEEE Trans.
  Geosci. Remote Sens.}, vol.~56, no.~6, pp. 3611--3623, 2018.

\bibitem{MendezDominguez2019}
E.~Mendez~Dominguez \emph{et~al.}, ``{A Back-Projection Tomographic Framework
  for VHR SAR Image Change Detection},'' \emph{IEEE Trans. Geosci. Remote
  Sens.}, vol.~57, no.~7, pp. 4470--4484, 2019.

\bibitem{Belloni2017}
C.~Belloni, N.~Aouf, T.~Merlet, and J.-M. Le~Caillec, ``{SAR Image Segmentation
  with GMMs},'' in \emph{Proc. IET Radar}, 2017, pp. 3--6.

\bibitem{Duan2018}
Y.~Duan \emph{et~al.}, ``{Adaptive Hierarchical Multinomial Latent Model with
  Hybrid Kernel Function for SAR Image Semantic Segmentation},'' \emph{IEEE
  Trans. Geosci. Remote Sens.}, vol.~56, no.~10, pp. 5997--6015, 2018.

\bibitem{Tschannen2016}
M.~Tschannen \emph{et~al.}, ``{Deep Structured Features for Semantic
  Segmentation},'' in \emph{Proc. IEEE EUSIPCO}, 2017.

\bibitem{Reyes2019}
M.~Fuentes~Reyes \emph{et~al.}, ``{SAR-to-Optical Image Translation Based on
  Conditional Generative Adversarial Networks—Optimization, Opportunities and
  Limits},'' \emph{Remote Sensing}, vol.~11, no.~17, 2019.

\bibitem{Wang2018d}
K.~Wang, G.~Zhang, Y.~Leng, and H.~Leung, ``{Synthetic Aperture Radar Image
  Generation With Deep Generative Models},'' \emph{IEEE Geosci. Remote. Sens.
  Lett.}, vol.~16, no.~6, pp. 912--916, 2018.

\bibitem{Liu2018c}
M.~Liu \emph{et~al.}, ``{Generating simulated SAR images using Generative
  Adversarial Network},'' in \emph{Proc. SPIE Applications of Digital Image
  Processing XLI}, 2018.

\bibitem{Yao2017}
W.~Yao, D.~Marmanis, and M.~Datcu, ``{Semantic Segmentation using Deep Neural
  Networks for SAR and Optical Image Pairs},'' in \emph{Proc. Big data from
  space}, 2017, pp. 2--5.

\bibitem{Shahzad2019}
M.~Shahzad \emph{et~al.}, ``{Buildings detection in VHR SAR images using fully
  convolution neural networks},'' \emph{IEEE Trans. Geosci. Remote Sens.},
  vol.~57, no.~2, pp. 1100--1116, 2019.

\bibitem{Long2015}
J.~Long, E.~Shelhamer, and T.~Darrell, ``{Fully Convolutional Networks for
  Semantic Segmentation},'' in \emph{Proc. IEEE CVPR}, 2015.

\bibitem{Henry2018}
C.~Henry, S.~M. Azimi, and N.~Merkle, ``{Road Segmentation in SAR Satellite
  Images With Deep Fully Convolutional Neural Networks},'' \emph{IEEE Geosci.
  Remote. Sens. Lett.}, vol.~15, no.~12, pp. 1867--1871, 2018.

\bibitem{Wu2019a}
W.~Wu \emph{et~al.}, ``{PolSAR Image Semantic Segmentation Based on Deep
  Transfer Learning—Realizing Smooth Classification With Small Training
  Sets},'' \emph{IEEE Geosci. Remote. Sens. Lett.}, vol.~16, no.~6, pp.
  977--981, 2019.

\bibitem{Wu2018b}
W.~Wu \emph{et~al.}, ``{High-Resolution PolSAR Scene Classification with
  Pretrained Deep Convnets and Manifold Polarimetric Parameters},'' \emph{IEEE
  Trans. Geosci. Remote Sens.}, vol.~56, no.~10, pp. 6159--6168, 2018.

\bibitem{Ronneberger2015}
O.~Ronneberger, P.~Fischer, and T.~Brox, ``{U-Net: Convolutional Networks for
  Biomedical Image Segmentation},'' in \emph{Proc. MICCAI}, vol. 9351.\hskip
  1em plus 0.5em minus 0.4em\relax Springer LNCS, 2015, pp. 234--241.

\bibitem{Moreira2013}
A.~Moreira \emph{et~al.}, ``{A tutorial on synthetic aperture radar},''
  \emph{IEEE Geoscience and Remote Sensing Magazine}, vol.~1, no.~1, pp. 6--43,
  2013.

\bibitem{Zlateski2018}
A.~Zlateski, R.~Jaroensri, P.~Sharma, and F.~Durand, ``{On the Importance of
  Label Quality for Semantic Segmentation},'' in \emph{Proc IEEE/CVF CVPR},
  2018, pp. 1479--1487.

\end{thebibliography}

\end{document}